\pdfoutput=1

\documentclass[11pt]{article}
\usepackage[dvipsnames]{xcolor}
\usepackage{float}

\usepackage[]{emnlp2021}

\usepackage{times}
\usepackage{latexsym}

\usepackage[T1]{fontenc}

\usepackage[utf8]{inputenc}

\usepackage{microtype}

\usepackage{times}
\usepackage{booktabs}
\usepackage{array}
\usepackage{latexsym}
\usepackage{verbatim}
\usepackage{arydshln}
\usepackage{enumitem}
\usepackage{pgfplots}
\pgfplotsset{compat=1.17}
\usepgfplotslibrary{units}
\pgfplotsset{unit code/.code 2 args={\si{#1#2}}}
\usetikzlibrary{patterns}
\usepackage{siunitx}
\usepackage{subfigure}

\usepackage{microtype}

\usepackage{soul}
\usepackage{enumitem}
\usepackage{float}

\usepackage{amsmath}

\DeclareMathOperator*{\argmin}{arg\,min}

\title{Efficient Explanations from Empirical Explainers}

\author{Robert Schwarzenberg$^1$ \\ \And Nils Feldhus$^1$ \\ $^1$German Research Center for Artificial Intelligence (DFKI) \\ $^2$Technische Universität Berlin (TU Berlin) \\
  \texttt{\{firstname.lastname\}@dfki.de} \And Sebastian Möller$^{1,2}$}

\pgfplotstableread[row sep=\\,col sep=&]{
samples & mse \\
1 & 0.02009747363626957 \\
2 & 0.0008429235895164311 \\
3 & 0.0028204787522554398 \\
4 & 0.0006463826866820455 \\
5 & 0.0008646327769383788 \\
6 & 0.0004980604280717671 \\
7 & 0.00038995585055090487 \\
8 & 0.00047371594700962305 \\
9 & 0.00022586091654375196 \\
10 & 0.0003254343173466623 \\
11 & 0.00013927304826211184 \\
12 & 0.00017835976905189455 \\
13 & 8.784554665908217e-05 \\
14 & 9.223689266946167e-05 \\
15 & 8.621902816230431e-05 \\
16 & 4.326146517996676e-05 \\
17 & 9.233197488356382e-05 \\
18 & 1.2237337614351418e-05 \\
19 & 8.6949672549963e-05 \\
}\bertimdbig

\def\empbertimdbig{0.0005650797393172979}

\pgfplotstableread[row sep=\\,col sep=&]{
samples & mse \\
1 & 0.11436866223812103 \\
2 & 0.060690347105264664 \\
3 & 0.0419205017387867 \\
4 & 0.03289978951215744 \\
5 & 0.02689073234796524 \\
6 & 0.02354772388935089 \\
7 & 0.02137296460568905 \\
8 & 0.01935291476547718 \\
9 & 0.017613686621189117 \\
10 & 0.016513383015990257 \\
11 & 0.015483000315725803 \\
12 & 0.014734419994056225 \\
13 & 0.014148347079753876 \\
14 & 0.013390099629759789 \\
15 & 0.01266426406800747 \\
16 & 0.012150374241173267 \\
17 & 0.01195993460714817 \\
18 & 0.0116551723331213 \\
19 & 0.011266272515058517 \\
}\xlnetsnlisvs

\def\mseempxlnetsnlisvs{0.015143373049795628}

\pgfplotstableread[row sep=\\,col sep=&]{
samples & mse \\
1 & 0.08521036058664322 \\
2 & 0.02252725139260292 \\
3 & 0.019968003034591675 \\
4 & 0.13376593589782715 \\
5 & 0.012703930959105492 \\
6 & 0.04719444736838341 \\
7 & 0.01296765636652708 \\
8 & 0.02077641524374485 \\
9 & 0.015887055546045303 \\
10 & 0.0146497106179595 \\
11 & 0.026627417653799057 \\
12 & 0.03784976527094841 \\
13 & 0.011770923621952534 \\
14 & 0.017773833125829697 \\
15 & 0.013477809727191925 \\
16 & 0.013396737165749073 \\
17 & 0.020373111590743065 \\
18 & 0.008601921610534191 \\
19 & 0.014985114336013794 \\
20 & 0.011742949485778809 \\
21 & 0.013403679244220257 \\
22 & 0.011743352748453617 \\
23 & 0.008417577482759953 \\
24 & 0.011827594600617886 \\
}\agrobertaig

\def\empagrobertaig{0.008950088173151016}

\pgfplotstableread[row sep=\\,col sep=&]{
samples & mse \\
1 & 0.0593469962477684 \\
2 & 0.03138568624854088 \\
3 & 0.02196415141224861 \\
4 & 0.017068399116396904 \\
5 & 0.014222243800759315 \\
6 & 0.012271391227841377 \\
7 & 0.010963850654661655 \\
8 & 0.009948823601007462 \\
9 & 0.009151404723525047 \\
10 & 0.008526419289410114 \\
11 & 0.008069692179560661 \\
12 & 0.007565542124211788 \\
13 & 0.007194985169917345 \\
14 & 0.006926081608980894 \\
15 & 0.00665921438485384 \\
16 & 0.006417832802981138 \\
17 & 0.006180962081998587 \\
18 & 0.005989693105220795 \\
19 & 0.005870121065527201 \\
}\pawselectrasvs

\def\mseemppawselectrasvs{0.004497139248996973}

\begin{document}
\raggedbottom
\maketitle
\begin{abstract}
Amid a discussion about Green AI in which we see explainability neglected, we explore the possibility to efficiently approximate computationally expensive explainers. To this end, we propose feature attribution modelling with Empirical Explainers. Empirical Explainers learn from data to predict the attribution maps of expensive explainers. We train and test Empirical Explainers in the language domain and find that they model their expensive counterparts surprisingly well, at a fraction of the cost. They could thus mitigate the computational burden of neural explanations significantly, in applications that tolerate an approximation error. 
\end{abstract}
\section{Introduction}

In recent years, important works were published on the ecological impacts of artificial intelligence and deep learning in particular, e.g.~\citet{strubell2019energy}, \citet{schwartz2019green},  \citet{Henderson2020TowardsTS}. Research is focused on the energy hunger of model training and subsequent inference in production. Besides training and in-production inference, explainability has become an integral phase of many neural systems. 

In the ongoing discussion about Green AI we see explainability neglected. 
Conversely, in the explainability community, even though research on efficiency is an active area, apparently the discussion is currently shaped by other aspects, such as faithfulness and plausibility \cite{Jacovi2020TowardsFI}. 
This is surprising because to explain a single model output, many prominent explanation methods, in particular many feature attribution methods (cf. below), require a multiple of computing power when compared to the prediction step. 
\subsection{Motivation: Expensive Explainers}

\begin{figure*}[!htb]
    \centering
    \includegraphics[width=\textwidth]{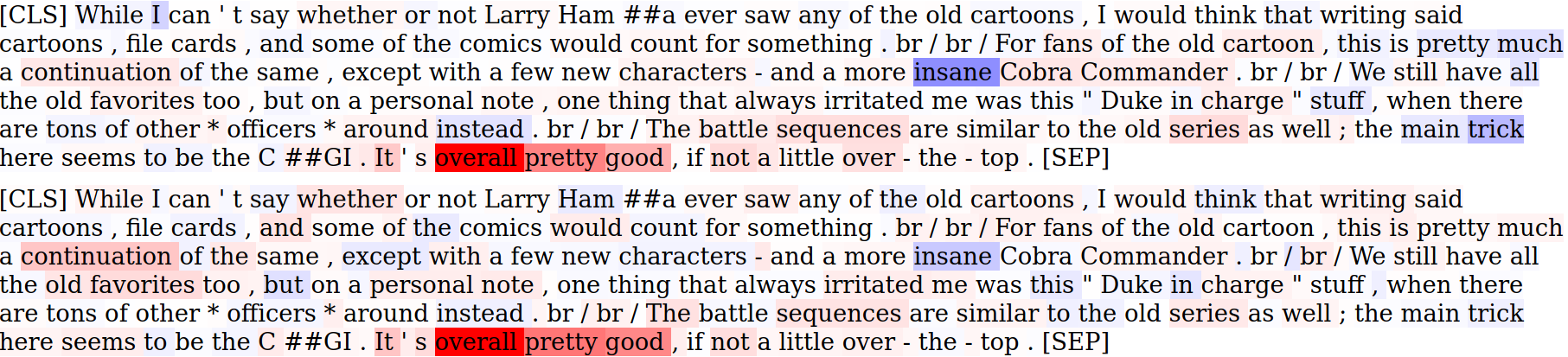}
    \caption{Explanations (attribution maps) for a BERT-based sentiment classification (best viewed digitally). The input is taken from the test split and was classified into \texttt{Positive}. \textbf{Top}: Integrated Gradients ($s=20$, $40$ passes through classifier required). \textbf{Bottom}: Empirical Integrated Gradients (1 pass through Empirical Explainer required). Attribution scores were normalized on sequence level. Red: positive; blue: negative.}
    \label{fig:heatmap-imdb-positive}
\end{figure*}
Take, for instance, the demonstrative but arguably realistic case of a classifier that was trained on $100k$ instances for 10 epochs. The training thus amounts to at least $1M$ forward passes and $1M$ backward passes. To produce explanations, in this paper, we consider feature attribution methods and focus on Integrated Gradients (IG) \cite{sundararajan2017axiomatic} and Shapley Values (SV) \cite{castro2009polynomial}, which are popular and established but also computationally expensive. To compute the exact IG or SV is virtually intractable, which is why sampling-based approximations were devised. For IG
\begin{equation*}
\label{eq:ig}
\phi_{f,i}(x) = \frac{x_{i} - \bar{x}_{i}}{s}\sum_{k=1}^{s} \frac{\partial f(\bar{x} + \frac{k}{s}(x - \bar{x}))}{\partial x_{i}}
\end{equation*}
is computed, where $x$ is the input to model $f$, $\bar{x}$ is a user-defined baseline, $s$ denotes the number of samples (a hyperparameter), and $\phi_{f,i}(x)$ denotes the attribution score of feature $i$. For SV, $s$ permutations of the input data $O_{1}, O_{2}, \dots O_{s}$ are drawn and then features from $x$ are added to a user-defined baseline,\footnote{There are several variants of Shapley Value Sampling. This sampling method is based on PyTorch's Captum \cite{kokhlikyan2020captum} library, that we also use for our experiments: \url{https://captum.ai/api/shapley_value_sampling.html}, last accessed March 26, 2021.} in the order they occur in the permutation. Let Pre$^{i}(O)$ denote the baseline including the features that were added to the baseline prior to $i$. The Shapley value can then be approximated by   
\begin{equation*}
    \phi_{f,i}(x) =\frac{1}{s}\sum_{k=1}^{s}f(\text{Pre}^{i}(O_{k})\cup x_{i}) - f(\text{Pre}^{i}(O_{k}))
\end{equation*} 

\citet{sundararajan2017axiomatic} report that $s$ between $20$ and $300$ is usually enough to approximate IG. Let us set $s:=20$. This requires 40 passes (forward and backward) through model $f$ to explain a single instance in production and furthermore, after only 50k explanations the computational costs of training are also already surpassed. In the case of SV, again setting $s:=20$ and assuming only $512$ input features (i.e. tokens to an NLP model), one already needs to conduct $20 * 512 = 10240$ passes to generate an input attribution map for a single classification decision.
This means that SV surpasses the training costs specified above after only $195$ explanations. 

This may only have a small impact if the number of required explanations is low. However, there are strong indications that explainability will take (or retain) an important role in many neural systems: For example, there are legal regulations, such as
the EU's GDPR which hints at a ``right to explanation'' \cite{goodman2017european}. For such cases, a 1:1 ratio in production between model outputs and explanations is not improbable.
If the employed explainability method requires more than one additional pass through the model (as many do, cf. below), there then is a tipping point at which the energy need of explanations exceeds the energy needs of both model training and in-production inference. 

IG and SV are not the only tipping point methods. Other expensive prominent and recent methods and variants are proposed by \citet{Zeiler2014VisualizingAU, ribeiro2016should, lundberg2017unified, smilkov2017smoothgrad, chen2018shapley, dhamdhere2019shapley, erion2019learning, covert2020improving, schwarzenberg2020pattern, schulz2020restricting, Harbecke2020ConsideringLI}. 

All of the above listed explainers require more than one additional pass through the model. This is why in general the following
should hold across methods: \textit{The smaller the model, the greener the explanation.} In terms of energy efficiency, explainability therefore benefits from model compression, distillation, or quantization. These are dynamic fields with a lot of active research which is why in the remainder of this paper we instead focus on something else: The mitigation of the ecological impact of tipping-point methods that dominate the cost term in the example cited in this section. 

These are our main contributions in this paper:  
\begin{enumerate}[noitemsep]
    \item We propose to utilize the task of feature attribution modelling to efficiently model the attribution maps of expensive explainers.  
    \item We address feature attribution modelling with trainable explainers that we coin Empirical Explainers. 
    \item We evaluate their performance qualitatively and quantitatively in the language domain and establish them as an efficient alternative to computationally expensive explainers in applications where an approximation error is tolerable.  
\end{enumerate}
\begin{figure*}[!htb]
    \centering
    \includegraphics[width=\textwidth]{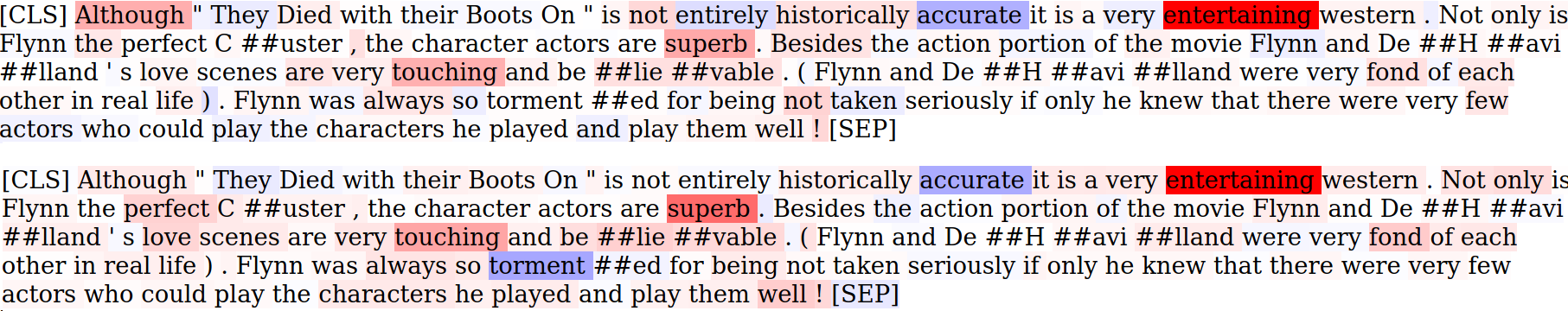}
    \caption{Explanations (attribution maps) for a BERT-based sentiment classification (best viewed digitally) with prominent approximation errors. The input is taken from the test split and was classified into \texttt{Positive}. \textbf{Top}: Integrated Gradients ($s=20$, $40$ passes through classifier required). \textbf{Bottom}: Empirical Integrated Gradients (1 pass through Empirical Explainer required). Attribution scores were normalized on sequence level. Red: positive; blue: negative. Note that contrary to the target explanation (top) the empirical integrated gradients for the token \texttt{tormented} are prominently negative (bottom).}
    \label{fig:heatmap-imdb-negative}
\end{figure*}
\section{Framework: Empirical Explainers}
\textit{Informally, an \textsc{empirical explainer} is a model that has learned from data to efficiently model the feature attribution maps of an expensive explainer. For training, one collects sufficiently many attribution maps from the expensive explainer and then maximizes the likelihood of these target attributions under the Empirical Explainer.} 

An expensive explainer may, for instance, be a costly attribution method such as Integrated Gradients that is used to return attributions for the decisions of a classifier, say, a BERT-based \cite{Devlin2019BERTPO}
sentiment classifier. The corresponding Empirical Explainer could be a separate neural network, similar in size to the sentiment classifier, consuming the same input tokens as the sentiment model, but instead of predicting the sentiment class, it is trained to predict the integrated gradients for each token.

Whereas the original Integrated Gradients explainer requires multiple passes through the classifier, producing the empirical integrated gradients 
requires just one pass through the similarly sized Empirical Explainer.  
Empirical explanations come with an accuracy-efficiency trade-off that we discuss in the course of a more formal definition of Empirical Explainers. 

For the more formal definition, we need to fix notation first. Let $E_{f}: {\rm I\!R}^{d} \rightarrow {\rm I\!R}^{d}$ be the expensive explainer that maps inputs onto attributions. 
Furthermore, let an Empirical Explainer be a function $e_{\theta}:  {\rm I\!R}^{d} \rightarrow {\rm I\!R}^{d}$, parametrized by $\theta$, which also returns attribution maps. Let $||\cdot||$ be a penalty for the inefficiency of a computation, e.g.~a count of floating point operations, energy consumption or number of model passes needed. Furthermore, let us assume, without the loss of generality, that $||E_{f}(x)|| >= ||e_{\theta}(x)||$ always holds; i.e.,~the Empirical Explainer -- which we develop and train -- is 
never more inefficient than the original, expensive explainer. Let $D: {\rm I\!R}^{d} \times {\rm I\!R}^{d} \rightarrow [0,1]$ be a similarity measure, where $D(l,m) = 0$ if $l=m$, for $l,m \in {\rm I\!R}^{d}$ and $\alpha, \beta \in [0,1]$ with $\alpha + \beta = 1$. For data $\mathbf{X}$, we define an $\alpha$-optimal Empirical Explainer by the $\argmin_{\theta \in \Theta}$
\begin{align}
\label{eq:maxlike}
\frac{1}{|X|}\sum_{x\in\mathbf{X}}\overbrace{\alpha D(E_{f}(x),e_{\theta}(x))}^{\text{accuracy}} + \overbrace{\beta \left(\frac{||e_{\theta}(x)||}{||E_{f}(x)||}\right) }^{\text{ efficiency}}. 
\end{align}

\subsection{Properties}
The first term describes how accurately the Empirical Explainer $e_{\theta}$ models the expensive explainer $E_{f}$. The second term compares the efficiency of the two explainers. For $\alpha = 1$, efficiency is considered unimportant and $e_{\theta} := E_{f}$ can be set to minimize Eq.~\ref{eq:maxlike}. $\alpha < 1$ allows to optimize efficiency at the cost of accuracy, which brings about the trade off: One may not succeed in increasing 
efficiency while maintaining accuracy. 
In fact, there is generally no exact guarantee for how accurately $e_{\theta}$ models $E_{f}$ for new data.  

Furthermore, while several expensive explainers, such as Integrated Gradients or Shapley Values, were developed axiomatically to have desirable properties, Empirical Explainers are derived from data -- empirically. Consequently, the evidence and guarantees Empirical Explainers offer for their faithfulness to the downstream model are empirical in nature and upper-bound by the faithfulness of the expensive explainer used to train them. 

We point this out explicitly because we would like to emphasize that we do \textit{not} regard an Empirical Explainer a new explainability method, nor do we argue that it can be used to replace the original expensive explainer everywhere. There are certainly situations for which Empirical Explainers are unsuitable for any $\alpha \neq 1$; critical cases in which explanations must have guaranteed properties. 

Nevertheless, we still see a huge potential for Empirical Explainers where approximation errors are tolerable: Consider, for instance, a search engine powered by a neural model in the back-end. Without the need to employ the expensive explainer, Empirical Explainers can efficiently provide the user with clues about what the model probably considers relevant in their query (according to the expensive explainer). 

\section{Experiments}
\label{sec:experiments}
\begin{figure*}[!htb]
    \centering
    \includegraphics[width=.8\textwidth]{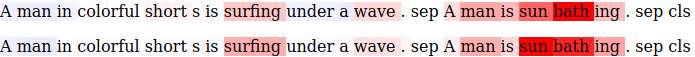}
    \caption{Explanations (attribution maps) for an XLNet-based NLI classification (best viewed digitally). The input is taken from the test split and was classified into \texttt{Contradiction}. \textbf{Top}: Shapley Value Samples ($s=20$, $380$ passes through classifier required). \textbf{Bottom}: Empirical Shapley Values (1 pass through Empirical Explainer required). Attribution scores were normalized on sequence level. Red: positive; blue: negative.}
    \label{fig:heatmap-snli-positive}
\end{figure*}

In this section, we report on the performance of Empirical Explainers that we trained and tested in the language domain.
We conducted tests with two prominent and expensive explainers, Integrated Gradients and Shapley Value Samples, varying the experiments across four state-of-the-art language classifiers, trained on four different tasks.  

\begin{figure*}[!htb]
    \centering
    \includegraphics[width=\textwidth]{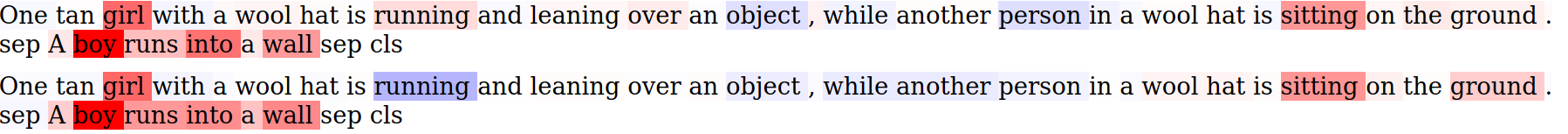}
    \caption{Explanations (attribution maps) for an XLNet-based NLI classification with prominent approximation errors (best viewed digitally). The input is taken from the test split and was classified into \texttt{Contradiction}. \textbf{Top}: Shapley Value Samples ($s=20$, $700$ passes through classifier required). \textbf{Bottom}: Empirical Shapley Values (1 pass through Empirical Explainer required). Attribution scores were normalized on sequence level. Red: positive; blue: negative. Note that, contrary to the target explanation (top), the empirical Shapley Value for the token \texttt{running} is in the negative regime (bottom).}
    \label{fig:heatmap-snli-negative}
\end{figure*}

The experiments address the question of whether or not it is feasible -- in principle -- to train efficient Empirical Explainers while achieving significant accuracy. All experiments, code, models and data are open source and can be retrieved following \url{https://github.com/DFKI-NLP/emp-exp}.
The most important choices are documented in the following paragraphs. Before going into greater detail, it is noteworthy that Eq.~\ref{eq:maxlike} provides a theoretical framework which one does not have use directly for explicit optimization. For example, in this work, we address the first objective, accuracy, by fitting Empirical Explainers to the attribution maps of expensive explainers. However, the second objective, efficiency, is addressed implicitly by design,~i.e.~the Empirical Explainers, in contrast to their expensive counterparts, are designed (and trained) in a way s.t.~only a single forward pass is required through a model similar in size to the downstream model. In the future, it would be very interesting to fully incorporate Eq.~\ref{eq:maxlike} in a differential setting, i.e.~also optimize for efficiency automatically, rather than by manual design. 

We trained four Empirical Explainers. All explainers consume only the input tokens to the downstream model and return an attribution score for each token. The first one (EmpExp-BERT-IG) was trained to predict integrated gradients w.r.t.~the input tokens to a BERT-based IMDB movie review  \cite{Maas2011LearningWV} 
classifier. For the second Empirical Explainer (EmpExp-XLNet-SV), we varied the downstream model architecture, task and target explainer: EmpExp-XLNet-SV predicts the Shapley Values (as returned by the expensive Shapley Value Sampling explainer) for the inputs of an XLNet-based \cite{Yang2019XLNetGA} natural language inference classifier that was trained on the SNLI \cite{Bowman2015ALA} 
dataset. The third (EmpExp-RoBERTa-IG) and fourth (EmpExp-ELECTRA-SV) empirical explainers again approximate IG and SV, but for a RoBERTa-based news topic classifier \cite{liu2019roberta} trained on the AG News dataset \cite{zhang2015character} and an ELECTRA (small)-based model \cite{clark2020electra} that detects paraphrases, trained on the PAWS dataset (subset ``labelled\_final'') \cite{zhang2019paws}, respectively. 

The Empirical Explainers were trained on target attributions that we generated with IG and SV with $s:=20$ samples. For EmpExp-RoBERTa-IG, we used $s:=25$ due to a slower convergence rate (cf.~below). Explanations were generated for the output neuron with the maximal activation. EmpExp-BERT-IG was trained with early stopping using the IG attribution maps for the full IMDB train split (25k). EmpExp-XLNet-SV was trained with around 100k SV attribution maps for the SNLI train split with early stopping, for which we used the 10k attribution maps for the validation split. We did not use all training instances in the split for EmpExp-XLNet-SV, due to the computational costs of Shapley Value Sampling. In case of EmpExp-RoBERTa-IG and EmpExp-ELECTRA-SV it was possible to train with the full train splits again, 108k (12k held out) and around 50k instances, respectively. 

As mentioned above, the expensive explainers require user-defined baselines. For the baselines, we replaced all non-special tokens in the input sequence with pad tokens. For the expensive IG, we produced attribution maps for the embedding layer and projected the attribution scores onto tokens by summing them over the token dimension. For the expensive SV, the input IDs were perturbed. During perturbation, we grouped and treated special tokens (CLS, SEP, PAD, ...) in the original input as one feature to accelerate the computation.

In architectural terms, the Empirical Explainers are very similar to the downstream models: We heuristically decided to copy the fine-tuned BERT, XLNet, RoBERTa and ELECTRA encoders from the classifiers and instead of the classification layers on top, we initialized new fully connected layers with $T$ output neurons. $T$ was lower bound by the maximum input token sequence length to the downstream model in the respective dataset: $T=512$ for BERT/IMDB and RoBERTa/AG News, $T=130$ for XLNet/SNLI, and $T=145$ for ELECTRA/PAWS. All input sequences were padded to $T$ and we did not treat padding tokens different from other tokens, when training the Empirical Explainers. Please note that the sequence length has a considerable impact on the runtime of the SV explainer in particular, which is why limiting $T$ increases comparability.

We trained the Empirical Explainers to output the right (in accordance with the expensive explainers) attribution scores for the input tokens, using an MSE loss between $E_{f}(x_{1})\dots E_{f}(x_{T})$ and $e_{\theta}(x_{1})\dots e_{\theta}(x_{T})$
where $x = x_{1}, x_{2}, \dots x_{T}$ is a sequence of input tokens.\footnote{Even though we do not solve for Eq.~\ref{eq:maxlike} directly, please note that for evaluation we can normalize the attribution scores to $[0,1]$ prior to computing the MSE and this way force the MSE into the interval $[0,1]$ to comply with the constraints for Eq.~\ref{eq:maxlike}.}

To put the performance of an Empirical Explainer into perspective, we propose the following baseline, which is the strongest we can think of: We take the original expensive explainer with a reduced number of samples as the baseline. To position the Empirical Explainer against this alternative energy saving strategy, we compute convergence curves. Starting with $s=1$, we incrementally increase the number of samples until $s=19$ ($s=24$ in the case of EmpExp-RoBERTa-IG) and collect attribution maps from the expensive explainer for the different choices of $s$. We then compute the MSEs of these attribution maps when compared to the target attributions (with $s=20$ or $s=25$). We average the MSEs across the test split. The same is done for the Empirical Explainer.
\section{Results \& Discussion}\label{sec:results}
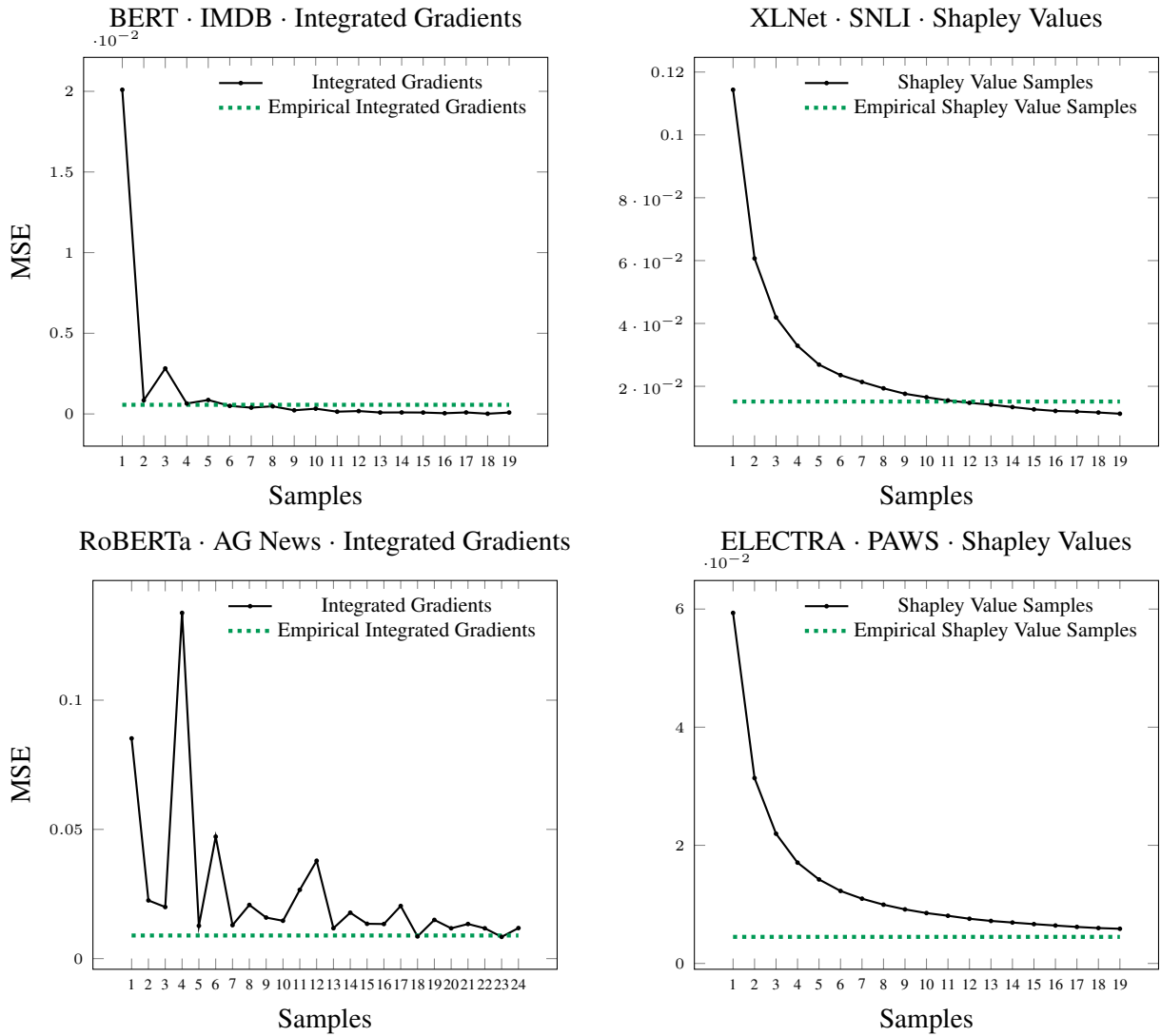
\begin{figure*}[!htb]
\begin{tikzpicture}
    \begin{axis}[
            title={BERT $\cdot$ IMDB $\cdot$ Integrated Gradients}, 
            height=7cm,  
            width=\textwidth/2,
            label style={font=\tiny},
            tick label style={font=\tiny}, 
            symbolic x coords={1,2,3,4,5,6,7,8,9,10,11,12,13,14,15,16,17,18,19, 20},
            xtick=data,
            xlabel={Samples}, 
            ylabel={MSE}, 
            mark size=.5pt, 
            legend style={draw=none, nodes={scale=0.75, transform shape}}, 
            ylabel near ticks, 
            xlabel near ticks, 
        ]
        \addplot[mark=*, thick] table[x=samples,y=mse]{\bertimdbig};
        \addlegendentry{Integrated Gradients}; 
        \addplot[dotted, ultra thick, smooth, color=ForestGreen] coordinates {(1,\empbertimdbig) (2,\empbertimdbig) (3,\empbertimdbig) (4,\empbertimdbig) (5,\empbertimdbig) (6,\empbertimdbig) (7,\empbertimdbig) (8,\empbertimdbig) (9,\empbertimdbig) (10,\empbertimdbig) (11,\empbertimdbig) (12,\empbertimdbig) (13,\empbertimdbig) (14,\empbertimdbig) (15,\empbertimdbig) (16,\empbertimdbig) (17,\empbertimdbig) (18,\empbertimdbig) (19,\empbertimdbig)};
        \addlegendentry{Empirical Integrated Gradients}; 
    \end{axis}
\end{tikzpicture}\hfill
\begin{tikzpicture}
    \begin{axis}[
            title={XLNet $\cdot$ SNLI $\cdot$ Shapley Values}, 
            height=7cm,  
            width=\textwidth/2,
            label style={font=\tiny},
            tick label style={font=\tiny}, 
            symbolic x coords={1,2,3,4,5,6,7,8,9,10,11,12,13,14,15,16,17,18,19},
            xtick=data,
            xlabel={Samples}, 
            mark size=.5pt, 
            legend style={draw=none, nodes={scale=0.75, transform shape}}, 
            ylabel near ticks, 
            xlabel near ticks, 
        ]
        \addplot[mark=*, thick] table[x=samples,y=mse]{\xlnetsnlisvs};
        \addlegendentry{Shapley Value Samples}; 
        \addplot[dotted, ultra thick, smooth, color=ForestGreen] coordinates {(1,\mseempxlnetsnlisvs) (2,\mseempxlnetsnlisvs) (3,\mseempxlnetsnlisvs) (4,\mseempxlnetsnlisvs) (5,\mseempxlnetsnlisvs) (6,\mseempxlnetsnlisvs) (7,\mseempxlnetsnlisvs) (8,\mseempxlnetsnlisvs) (9,\mseempxlnetsnlisvs) (10,\mseempxlnetsnlisvs) (11,\mseempxlnetsnlisvs) (12,\mseempxlnetsnlisvs) (13,\mseempxlnetsnlisvs) (14,\mseempxlnetsnlisvs) (15,\mseempxlnetsnlisvs) (16,\mseempxlnetsnlisvs) (17,\mseempxlnetsnlisvs) (18,\mseempxlnetsnlisvs) (19,\mseempxlnetsnlisvs)};
        \addlegendentry{Empirical Shapley Value Samples}; 
    \end{axis}
\end{tikzpicture} \\
\begin{tikzpicture}
    \begin{axis}[
            title={RoBERTa $\cdot$ AG News $\cdot$ Integrated Gradients}, 
            height=7cm,  
            width=\textwidth/2,
            label style={font=\tiny},
            tick label style={font=\tiny}, 
            symbolic x coords={1,2,3,4,5,6,7,8,9,10,11,12,13,14,15,16,17,18,19, 20,21,22,23,24},
            xtick=data,
            xlabel={Samples}, 
            ylabel={MSE}, 
            mark size=.5pt, 
            legend style={draw=none, nodes={scale=0.75, transform shape}}, 
            ylabel near ticks, 
            xlabel near ticks, 
            scaled ticks=false, 
            tick label style={/pgf/number format/fixed} 
        ]
        \addplot[mark=*, thick] table[x=samples,y=mse]{\agrobertaig};
        \addlegendentry{Integrated Gradients}; 
        \addplot[dotted, ultra thick, smooth, color=ForestGreen] coordinates {(1,\empagrobertaig) (2,\empagrobertaig) (3,\empagrobertaig) (4,\empagrobertaig) (5,\empagrobertaig) (6,\empagrobertaig) (7,\empagrobertaig) (8,\empagrobertaig) (9,\empagrobertaig) (10,\empagrobertaig) (11,\empagrobertaig) (12,\empagrobertaig) (13,\empagrobertaig) (14,\empagrobertaig) (15,\empagrobertaig) (16,\empagrobertaig) (17,\empagrobertaig) (18,\empagrobertaig) (19,\empagrobertaig) (20,\empagrobertaig) (21,\empagrobertaig) (22,\empagrobertaig) (23,\empagrobertaig) (24,\empagrobertaig)};
        \addlegendentry{Empirical Integrated Gradients}; 
    \end{axis}
\end{tikzpicture}\hfill
\begin{tikzpicture}
    \begin{axis}[
            title={ELECTRA $\cdot$ PAWS $\cdot$ Shapley Values}, 
            height=7cm,  
            width=\textwidth/2,
            label style={font=\tiny},
            tick label style={font=\tiny}, 
            symbolic x coords={1,2,3,4,5,6,7,8,9,10,11,12,13,14,15,16,17,18,19, 20,21,22,23,24},
            xtick=data,
            xlabel={Samples}, 
            mark size=.5pt, 
            legend style={draw=none, nodes={scale=0.75, transform shape}}, 
            ylabel near ticks, 
            xlabel near ticks, 
        ]
        \addplot[mark=*, thick] table[x=samples,y=mse]{\pawselectrasvs};
        \addlegendentry{Shapley Value Samples}; 
        \addplot[dotted, ultra thick, smooth, color=ForestGreen] coordinates {(1,\mseemppawselectrasvs) (2,\mseemppawselectrasvs) (3,\mseemppawselectrasvs) (4,\mseemppawselectrasvs) (5,\mseemppawselectrasvs) (6,\mseemppawselectrasvs) (7,\mseemppawselectrasvs) (8,\mseemppawselectrasvs) (9,\mseemppawselectrasvs) (10,\mseemppawselectrasvs) (11,\mseemppawselectrasvs) (12,\mseemppawselectrasvs) (13,\mseemppawselectrasvs) (14,\mseemppawselectrasvs) (15,\mseemppawselectrasvs) (16,\mseemppawselectrasvs) (17,\mseemppawselectrasvs) (18,\mseemppawselectrasvs) (19,\mseemppawselectrasvs)};
        \addlegendentry{Empirical Shapley Value Samples};
    \end{axis}
\end{tikzpicture}
\caption{Performance of Empirical Explainers (dashed green lines) and convergence curves of expensive explainers  (solid black lines), averaged across test sets. The attribution maps returned by the expensive explainers with $s=20$ samples in case of BERT, XLNet and ELECTRA and $s=25$ in case of RoBERTa (slower convergence behaviour), were regarded the target explanations. MSEs were computed on a per-sequence basis and then averaged across the test set.
}
\label{fig:convergence-curves}
\end{figure*}
In the following, we report the experimental results, divided into the aspects of task performance, explanation efficiency and explanation accuracy.

\noindent
\textit{Task Performance}\quad On the test splits, the classifiers we trained achieved  weighted F$_{1}$ scores of 0.93 (BERT $\cdot$ IMDB), 0.90 (XLNet $\cdot$ SNLI), 0.94 (RoBERTa $\cdot$ AG News) and 0.92 (ELECTRA $\cdot$ PAWS). 

\noindent
\textit{Explanation Efficiency}\quad Regarding the efficiency term in Eq.~\ref{eq:maxlike}, in terms of model passes, the Empirical Explainers have a clear advantage over their expensive counterparts. For IG with $s=20$, 40 model passes are required, for $s=25$, 50 passes. For SV with $s=20$, assuming a token sequence length of 100 for the purpose of discussion, 2000 model passes are required. For the empirical explanations, only one (additional) forward pass through a similarly sized model (the Empirical Explainer) is necessary.    

Contrary to runtime (and energy consumption) measures, the number of required model passes is largely invariant of available hardware and implementation details. For the sake of completeness, we nevertheless also report our runtimes in appendix~\ref{app:runtimes}. In summary, generating the expensive explanations for the test splits took between around 02:15 and 48 \textit{hours}, whereas the empirical explanations only required between 02:05 and 07:14 \textit{minutes}.    

The runtimes are not definitive, however. We were unable to establish a fair game for the explainers. For example, due to implementation details and memory issues we explained the data instance-wise with the expensive explainers while our Empirical Explainers easily allowed batch processing. We expect that the expensive explainers can be accelerated but due to the larger number of model passes required, they will very likely not outperform their empirical counterparts. 

The Empirical Explainers come with additional training costs, which we also report in appendix \ref{app:runtimes}. Training took between 02:15 and 07:00 hours. These additional training costs are thus quickly outweighed by the expensive explainers, in particular in a continuous in-production setting.  

\noindent
\textit{Explanation Accuracy}\quad Regarding the accuracy term in Eq.~\ref{eq:maxlike}, Figs.~\ref{fig:heatmap-imdb-positive}, \ref{fig:heatmap-imdb-negative}, \ref{fig:heatmap-snli-positive} and \ref{fig:heatmap-snli-negative} provide anecdotal qualitative evidence that the Empirical Explainers are capable of modelling their expensive counterparts well, with varying degrees of approximation errors. Alongside this paper, we provide four files (see repository) with around 25k (IMDB), 10k (SNLI), 7.6k (AG News) and 8k (PAWS) lines, each of which contains an HTML document that depicts a target attribution and its empirical counterpart from the test set. The heatmaps in the figures mentioned above are taken from the accompanying files.

Figs.~\ref{fig:heatmap-imdb-positive} and \ref{fig:heatmap-snli-positive} are instances of what we consider surprisingly accurate approximations of the expensive target attribution maps, despite challenging inputs. Let us first consider Fig.~\ref{fig:heatmap-imdb-positive} in greater detail. Consider the tokens \texttt{favorites} and \texttt{irritated} that are not attributed much importance by the expensive explainer (IG, top) but could be considered signal words for the positive and negative class, respectively and thus pose a challenge for the Empirical Explainer. Nevertheless, in accordance with the expensive target explainer, the Empirical Explainer (bottom) does not attribute the classifier output primarily to these tokens but instead accurately assigns a lot of weight to \texttt{It's overall pretty good}. 

A similar phenomenon can be observed in Fig.~\ref{fig:heatmap-imdb-negative} for the token \texttt{love}. The approximation in this figure, however, also contains a prominent approximation error. The Empirical Explainer erroneously attributes a salient negative score to the token \texttt{tormented} while the target explainer does not highlight that token. Similarly, in Fig.~\ref{fig:heatmap-snli-negative} the Empirical Explainer returns a negative score for \texttt{running}, whereas the expensive target explainer has returned a positive score. 

We suspect that such errors may result from global priors that the Empirical Explainers have learned and that sometimes outweigh the instantaneous information. For instance, in Fig.~\ref{fig:heatmap-snli-negative} the verb \texttt{running} in the premise in conjunction with the (conjugated) verb \texttt{runs} in the hypothesis may statistically be indicative of an entailment in the training data. This is because to produce a contradiction the verb sometimes is simply replaced by another one (cf. Fig.~\ref{fig:heatmap-snli-positive}: \texttt{surfing} vs. \texttt{sun bathing}). In this instance, however, the verb is not replaced. Thus, here the prior knowledge of the Empirical Explainer may outweigh the local information in favor of the error that we observe: The Empirical Explainer may signal that \texttt{running} is evidence against the class \texttt{Contradiction} since it finds it in the premise and hypothesis. A similar argument can be put forward for the case of \texttt{tormented} in Fig.~\ref{fig:heatmap-imdb-negative}. 

The above points are rather speculative. A more objective and quantitative analysis of the efficiency/accuracy trade-off is provided in Fig.~\ref{fig:convergence-curves}. The left column depicts the MSE lines of IG for an increasing number of samples in $x$-direction. We observe that IG converges fast in case of BERT/IMDB. (This may be due to saturation effects in Integrated Gradients, reported on by \citet{miglani2020investigating}.) In case of RoBERTa/AG News we found a slower convergence rate, which is why we increased the number of samples for the target explainer. We observe that in both cases, the empirical integrated gradients (dashed lines) perform favourably: To outperform the Empirical Explainer by decreasing $s$, in case of BERT/IMDB one needs to set $s>5$ which entails $10$ model passes as opposed to the single additional pass through the Empirical Explainer for the empirical explanations.  Furthermore, the approximation error is already marginal at the intersection of expensive and empirical line. In case of RoBERTa/AG News, one even needs to set $s:=18$ to be closer to the target than the Empirical Explainer.  

A similar trend can be observed for the (empirical) Shapley Values in the right column of Fig.~\ref{fig:convergence-curves}. In case of XLNet/SNLI, however, the intersection occurs only after $s=10$ which means that the Empirical Explainer needs only $\frac{1}{11*100} = 0.9\%$ of model passes (plus the pass through its output layer) when compared to the next best expensive explainer, again assuming 100 input tokens for the purpose of discussion. In case of ELECTRA/PAWS, the Empirical Explainer even beats the expensive explainer with just one sample less than the target. 

In summary, we take the experimental results as a strong indication that Empirical Explainers could become an efficient alternative to expensive explainers (in the language domain) where approximation errors are tolerable.

\section{Related Work}
\label{sec:related-work}
The computational burden of individual explainability methods was addressed in numerous works. As mentioned above, Integrated Gradients can only be computed exactly in limit cases and for all other cases, the community relies on the approximate method proposed by \citet{sundararajan2017axiomatic}. Similarly, Shapley Values can rarely be computed precisely which is why Shapley Value Sampling was investigated, e.g. by  \citet{castro2009polynomial, trumbelj2010AnEE}. Shapley Value Sampling was later unified with other methods under the SHAP framework \citep{lundberg2017unified} which yielded the method KernelSHAP that showed improved sample efficiency. \citet{covert2020improving} then analysed the convergence behaviour of KernelSHAP and again further improved runtime. \citet{chen2018shapley} introduced L-Shapley and C-Shapley which accelerate Shapley Value Sampling for structured data, such as dependency trees in NLP. 

Thus, computational feasibility appears to be a driving force in the research community, already. To the best of our knowledge, however, we are the first to propose the task of feature attribution modelling to mitigate the computational burden of expensive explainers with Empirical Explainers. 

Technically, self-explaining models \citep{alvarez2018towards} are related to our approach in that they also generate explanations in a forward pass (alongside their classification decision). Contrary to self-explaining models, Empirical Explainers can be employed after training for a variety of black box and white box classifiers and explainers. 

A source of inspiration for our method was the work by \citet{camburu2018snli}. The authors train self-explaining models that return a natural language rationale alongside their classification. Thus, they, too, train an explainer. However, their target explanations (natural language) differ substantially from the ones Empirical Explainers are trained with (attribution scores). 

Furthermore, related to our work is the technique of gradient matching for which a network's (integrated) gradients are compared to a target attribution, i.e. a human prior, and then the network's parameters are updated, s.t. the gradients move closer to the target, as done e.g.~by \citet{ross2017right, erion2019learning, liu2019incorporating}. Apart from the loss on an alignment with target attributions, our method and goals diverge from theirs significantly. 

Human priors and expensive target explanations have recently also been used for \textit{explanatory interactive learning} (XIL). Like Empirical Explainers, XIL is motivated by the expensiveness of a target explainer; in the case of XIL this is a human in the loop. \citet{schramowski2020making} present humans with informative (cf. active learning) instances, the model prediction and an explanation for the prediction
The expensive human feedback is then used to improve the model. 
Apart from the expensive explainer assumption, their approach differs substantially from ours. Very recently, \citet{behrens2021bandits} contributed to XIL by introducing a method that learns to explain from explanations and in this respect is close to the setting of Empirical Explainers. One fundamental difference between ours and their work is that they propose and focus on a specific class of self-explainable models  whereas Empirical Explainers make no assumptions about the underlying predictor and are intended for a variety of model classes, as already mentioned above. 

Very recently again, \citet{Rajagopal2021SelfExplainAS} proposed local interpretable layers as a means to generate concept attributions which in parts aligns with our method, even though their target attributions and task objectives are very different again.

Lastly, Empirical Explainers can be viewed as a form of knowledge distillation \cite{hinton2015distilling}. However, contrary to the established approach, we do not assume a parametric teacher network that knowledge is distilled from. Very recently we became aware of the work by  
\citet{pruthi2020evaluating} who boost the accuracy of a student learner with explanations in the form of a subset of tokens that are relevant to the teacher decision, determined by an explainer. In a sequence classification task, the student is trained to identify the relevant tokens and could thus be considered an Empirical Explainer. The task, however, is not to predict the original attribution map and the overall objective differs significantly from ours again. 

\section{Conclusion \& Future Directions}
In this paper, we take a step towards greener XAI by again reviving energy efficiency as an additional criterion by which to judge an explainability method, alongside important aspects such as faithfulness and plausibility. In this context, we propose feature attribution modelling with efficient Empirical Explainers. In the language domain, we investigate the efficiency/accuracy trade-off and find that it is possible to generate empirical explanations with significant accuracy, at a fraction of the costs of the expensive counterparts. We take this as a strong indication that Empirical Explainers could be a viable alternative to expensive explainers where approximation errors are tolerable. 

Regarding future directions: The Empirical Explainers we trained are our concrete model choices. The framework we propose allows for many other approaches. For instance, one could provide the Empirical Explainers with additional information, such as the gradient w.r.t.~the inputs. This would require an additional pass through the model but may possibly further boost accuracy.  

We would like to note that we trained and tested our Empirical Explainers only on in-domain data but their behaviour on out-of-domain data should be investigated, too. Fortunately, since we explain the model decision (the maximum output activation), no gold labels are required to train Empirical Explainers which facilitates data collection immensely. Finally, there are some more sample efficient explainers that should be considered, too.

\paragraph{Acknowledgements}{\small We would like to thank Christoph Alt, Steffen Castle, Jonas Mikkelsen and David Harbecke for their helpful feedback. This work has been supported by the German Federal Ministry of Education and Research as part of the projects XAINES (01IW20005) and CORA4NLP (01IW20010).}

\bibliography{anthology,literature}
\bibliographystyle{acl_natbib}
\appendix
\section{Runtimes}
\label{app:runtimes}
Generating the expensive target explanations for the official IMDB test split (25k instances, $T=512$, $s=25$, BERT, Titan V) with Integrated Gradients took us 7:17 hours
(6:22 hours for the 22500 training instances). Generating the expensive Shapley Values for the SNLI test split ($\sim$ 10k instances, $T=130$, $s=20$, XLNet, Quadro P5000) took us 48:22 hours
(and over 600 GPU hours for under 100k training instances).
It took us over 2:15 hours to explain the 7600 instances in the test split of the AG News dataset with IG ($T=512$, $s=25$, RoBERTa, RTX2080Ti; over 31 hours for the train split). For the PAWS test split ($T=145$, $s=20$, 8k instances, ELECTRA (small), RTX6000) we needed over 18 GPU hours 
(over 126 GPU hours for the 49401 instances in the train split, using NVIDIA's RTX3090 and RTX2080Ti).  

In contrast, generating the empirical explanations took us only 07:14 \textit{minutes} for the IMDB test split on the Titan GPU and only
02:05 \textit{minutes} for SNLI test split on the Quadro P5000 GPU.  The AG News test split took 03:16 \textit{minutes} to explain (RTX3090) and the PAWS test split was explained empirically in only 02:19 \textit{minutes} (RTX3090).  

The training of EmpExp-BERT-IG terminated after 10 epochs (Titan V), which took less than 4 hours. EmpExp-XLNet-SV (Quadro P5000), EmpExp-RoBERTa-IG (RTX3090), and EmpExp-ELECTRA-SV (RTX3090) terminated after 7, 3, and 8 epochs, respectively (<7 hours, < 2:15 hours, and <1 hours).




\end{document}